\documentclass[10pt,twocolumn,letterpaper]{article}

\usepackage{iccv}
\usepackage{times}
\usepackage{epsfig}
\usepackage{graphicx}
\usepackage{amsmath}
\usepackage{amssymb}

\usepackage{microtype}
\usepackage{graphicx,caption}
\usepackage{subfigure}
\usepackage{booktabs} 
\usepackage{amsmath,amsfonts}
\usepackage{adjustbox}
\usepackage{xcolor}
\usepackage{booktabs}
\usepackage{algorithmic}
\usepackage{algorithm}

\graphicspath{{./imgs/}}

\usepackage[breaklinks=true,bookmarks=false]{hyperref}

\iccvfinalcopy 

\def\iccvPaperID{5621} 

\begin{document}

\title{Discriminative Clustering for Robust Unsupervised Domain Adaptation}

\author{Rui Wang \qquad Guoyin Wang \qquad Ricardo Henao \\
Duke University\\
{\tt\small rui.wang16, guoyin.wang, ricardo.henao@duke.edu}
}

\maketitle

\begin{abstract}
	Unsupervised domain adaptation seeks to learn an invariant and discriminative representation for an unlabeled target domain by leveraging the information of a labeled source dataset.
	We propose to improve the discriminative ability of the target domain representation by simultaneously learning tightly clustered target representations while encouraging that each cluster is assigned to a unique and different class from the source.
	This strategy alleviates the effects of negative transfer when combined with adversarial domain matching between source and target representations.
	Our approach is robust to differences in the source and target label distributions and thus applicable to both balanced and imbalanced domain adaptation tasks, and with a simple extension, it can also be used for partial domain adaptation.
	Experiments on several benchmark datasets for domain adaptation demonstrate that our approach can achieve state-of-the-art performance in all three scenarios, namely, balanced, imbalanced and partial domain adaptation.
\end{abstract}

\section{Introduction}
Deep neural networks have demonstrated remarkable advancement in supervised learning for a wide variety tasks in the past decade.
However, training such models usually requires the availability of massive labeled data, which is prohibitive in some applications.
Therefore, it is of interest to develop domain invariant classification models that are able to generalize to other (unlabeled) domains beyond that for which they were trained.
Unsupervised domain adaptation is a general framework for learning domain invariant representations.
The goal is to learn a shared latent representation (encoding) of (labeled) source and (unlabeled) target instances complemented with a classifier to accurately label instances using the latent representation as input.
During learning, the differences between source and target representations are minimized at a population (distribution) level, while the discriminative ability of the classifier is maximized using only the labeled source data.
Subsequently, the learned classifier and encoding can be used to label target instances without the need for manual labeling effort.

A typical application for this setting is image classification, where the instances are images, labels denote different image classes, the latent representations are often obtained via a convolutional encoder, and the source and target domains consist of instances of the same image classes but obtained under different technical conditions.

Provided there is no available labeled data for the target domain, existing unsupervised domain adaptation methods generally match the distributions of the source and target latent representations (features).
This approach assumes that the source and target share the same label domain and distribution, \emph{i.e.}, same labels and comparable label prevalences.
However, it cannot be guaranteed that the representation learned by distribution matching is discriminative, \emph{i.e.}, latent representations for different classes may not be well separated, thus difficult to classify.
Fortunately, this matching approach has been widely successful in practice, particularly, in image classification problems \cite{DAN, DANN, ADDA}.

\begin{figure}[t!]
	\centering
	\includegraphics[width=0.7\columnwidth]{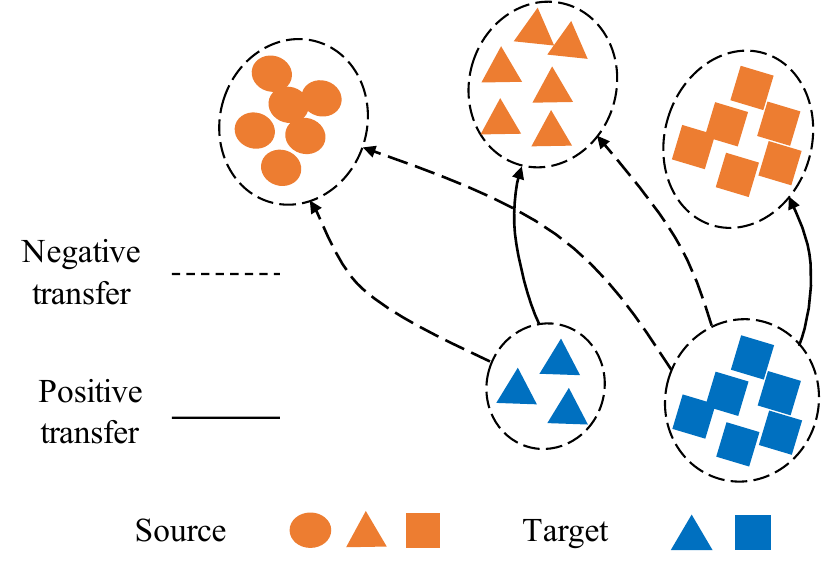}
	\caption{Negative transfer in partial domain adaptation. Source and target features are colored orange and blue, respectively. Instances from prevalent classes in the target will be negatively transferred to smaller classes or those not present in the target, which results in poorly discriminative target features.}
	\label{fg:negtrans}
	\vspace{-3mm}
\end{figure}

In real applications, the unknown target label distribution can be different from the source, \emph{i.e.}, labels in the target domain can be observed with different proportions compared to the source.
Further, in the case of \emph{partial domain adaptation} \cite{SAN,PADA}, the set of labels in the target domain can be a subset of the source.
As illustrated in Figure~\ref{fg:negtrans}, these scenarios are challenging because matching the distribution of the latent features for source and target domains is likely to result in \emph{negative transfer}.
This happens because distribution matching forces observations from the target to be placed nearby source observations whose label is not present in the target, thus negatively impacting the quality of the target encoder.
As a result, the adapted model may be sometimes worse than that trained on the source, since the target representation is poorly or lacks discriminative ability after adaptation.

In this paper, we propose an approach that extends the scope of unsupervised domain adaptation by relaxing the assumption of needing shared label domain and distributions.
Consequently, our approach is robust to differences in source and target label distributions.
Assuming that the source label distribution is uniform, which can be easily achieved by resampling, we consider the following three scenarios:

\textbf{Balanced domain adaptation}: This is the case when the target label distribution is balanced in relation to the source, which is
 the basic assumption of most previous methods \cite{DAN, DANN, ADDA}, and can be addressed by distribution matching of the source and target features.

\textbf{Imbalanced domain adaptation}: When the target label proportions are substantially different in relation to the source, thus \emph{imbalanced}.
In this scenario, negative transfer is likely to occur, which will result in degraded performance for some classes, usually the most prevalent ones, as illustrated in Figure~\ref{fg:negtrans}.
%

\textbf{Partial domain adaptation:} This scenario, recently studied in \cite{SAN,PADA} considers the case when the target label domain is a subset of the source.
In some sense, it can be understood as an extreme case of class imbalance, that in which some of the target classes occur with zero probability.
However, these two scenarios need to be considered separately because in terms of performance, they need to be evaluated differently.
For instance, in imbalanced domain adaptation the overall classification accuracy is not necessarily meaningful or informative.
Alternatively, one may consider class-wise accuracies which may be a more appropriate performance metric.
%

In this work, we propose a new approach that accommodates the three scenarios described above.
Our key contribution is to improve the discriminative ability of the target latent representation by simultaneously $i$) learning tightly clustered target representations, $ii$) encouraging that each cluster gets assigned to a different and unique class from the source, and $iii$) minimizing the discrepancies between source and target representations by distribution matching.
We will show empirically that these three criteria largely alleviate the effects of negative transfer on imbalanced and partial domain adaptation tasks.
Finally, our experiments demonstrate that our approach yields excellent results on all three scenarios.

\section{Robust Unsupervised Adaptation}
Our approach extends the ability of current domain matching adaptation models to the \emph{imbalanced} and \emph{partial} settings.
This is achieved by learning a tightly clustered target representation while encouraging that each cluster is assigned to a unique and different class from the source.
These two criteria are combined with representation distribution matching as in Adversarial Discriminative Domain Adaptation (ADDA) \cite{ADDA}, which will result in more discriminative and domain invariant target representations, as will be demonstrated in the experiments.

Assume we have a labeled \emph{source} dataset, $\{X_s,Y_s\}_{s=1}^{N_s}$, where $X_s$ and $Y_s$ represent source inputs and labels, respectively, and $N_s$ is the number of observed pairs.
Source labels, $Y_s\in{\cal Y}_s$, can take one of $K_s$ distinct labels with (marginal) probability $P(Y_s)$.
We seek to leverage information in the source and a set of (unlabeled) \emph{target} inputs, $\{X_t\}_{t=1}^{N_t}$ of size $N_t$, to make predictions about their labels, \emph{i.e.}, to obtain $Y_t$ for $X_t$.
Similarly, $Y_t\in{\cal Y}_t$ can take one of $K_t$ distinct labels with (marginal) probability $P(Y_t)$.
Here we not only consider the standard scenario, denoted as \emph{balanced domain adaptation}, where ${\cal Y}_s={\cal Y}_t$ and $P(Y_s)=P(Y_t)$, but also \emph{imbalanced domain adaptation}, where ${\cal Y}_s={\cal Y}_t$, but $P(Y_s)\neq P(Y_t)$, \emph{i.e.}, the set of labels in source and target domains are the same but observed in different proportions.
Further, we also consider \emph{partial domain adaptation}, where ${\cal Y}_t\subset{\cal Y}_s$, \emph{i.e.}, the target labels are a true subset of the source labels, so $K_t<K_s$.
This scenario can be seen as an extreme case of imbalanced domain adaptation where some labels in the target domain are observed with probability zero.

\begin{figure}[t]
	\centering
	\includegraphics[width=\columnwidth]{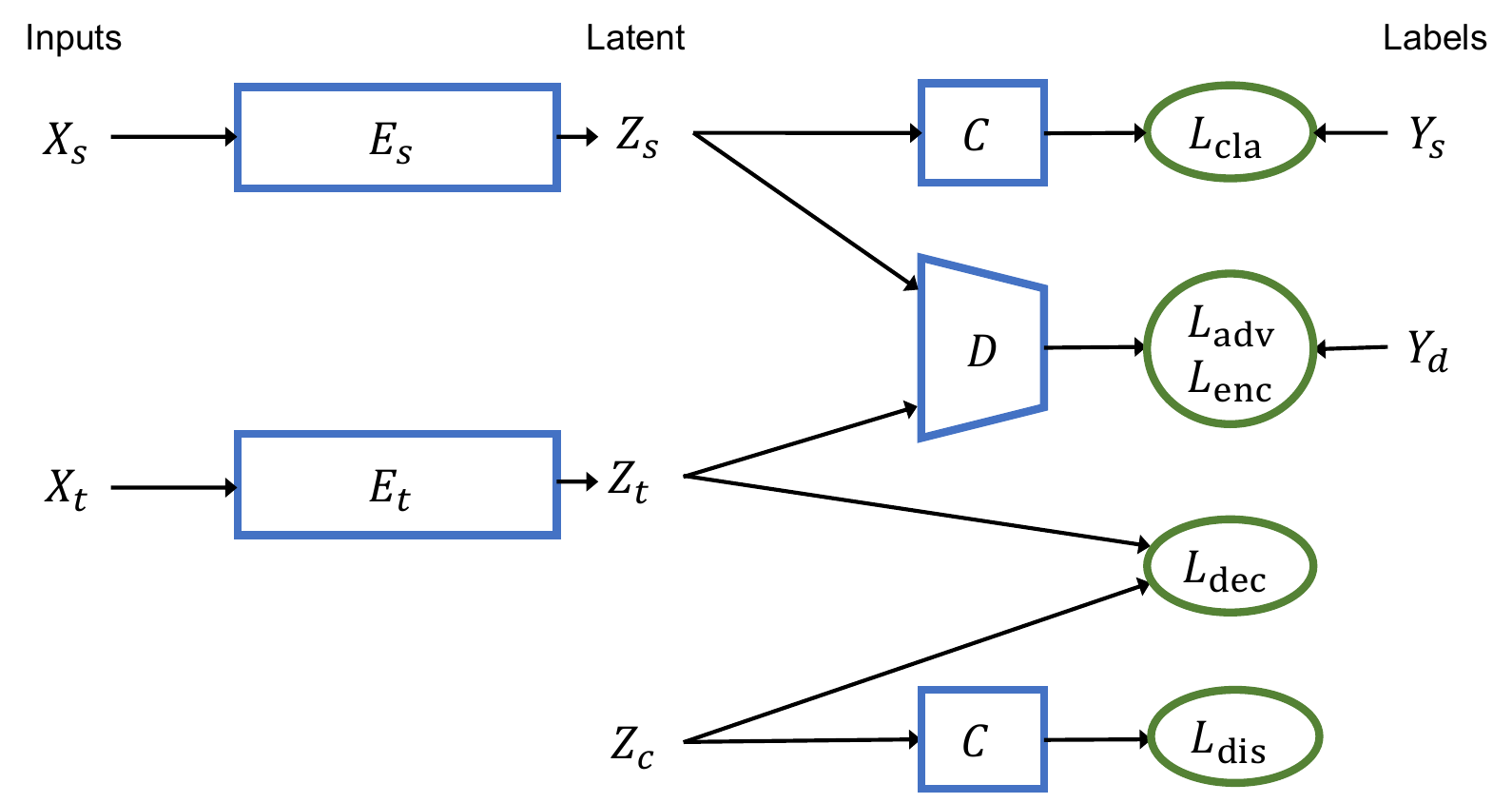}
	\caption{Robust unsupervised domain adaptation architecture. $Y_s$ and $Y_d$ are source and domain labels, respectively. Latent features for source, target and target cluster centroids are denoted as $Z_s$, $Z_t$ and $Z_c$, respectively. Model blocks are represented as rectangles and losses as ellipses.
	}
	\label{fg:arch}
	\vspace{-2mm}
\end{figure}

Our approach assumes that we can obtain a source encoder, $Z_s=E_s(X_s)$, and classifier, $p(Y_s=k|Z_s)=C(Z_s)$, for $k=1,\ldots,K_s$, that can be trained on $\{X_s,Y_s\}_{s=1}^{N_s}$.
To obtain latent features that are uninformative of the differences between source and target domains, thus in principle only containing information about the labels, we specify a discriminator, $p(Y_{domain}=s|Z_d)=D(Z_d)$, tasked to learn whether $Z_d$, for $d\in\{s,t\}$, is from the source or target, $Y_{domain}\in\{s,t\}$.
This means that we seek to learn a discriminator such that $p(Y_{domain}=s|Z_s)=D(Z_s)\to 1$ and $p(Y_{domain}=s|Z_t)=D(Z_t)\to 0$.
This is done in an adversarial fashion, similar to ADDA \cite{ADDA}.
Further, we encourage the target latent representation, $Z_t$, to be clustered into $K_s$ components with centroids, $Z_c$, for $c=1\ldots,K_s$.
The objective for the model illustrated in Figure~\ref{fg:arch} consists of five terms, namely, classification, ${L_{\rm cla}}$, adversarial, $L_{\rm adv}$, encoder, $L_{\rm enc}$, clustering $L_{\rm dec}$, and dissimilarity, $L_{\rm dis}$, which we describe below.

%
%

\subsection{Classification objective}
The supervised model consists of the source encoder $Z_s=E_s(X_s)$ and classifier $p(Y_s=k|Z_s)=C(Z_s)$, for $k=1,\ldots,K_s$.
These are pre-trained on $\{X_s,Y_s\}_{s=1}^{N_s}$ by maximizing the following objective
%
\begin{equation}\label{eq:l_cla}
	L_{\rm cla} = {\mathbb E}_{(x, y)\sim p(X_s, Y_s)} [y^\top\log\{ C(E_s(x)) \}] \,,
\end{equation}
where $y$ is $K_s$-dimensional \emph{one-hot-vector} representation for $Y_s$, $C(Z_s)$ is assumed to have a softmax activation function and $p(X_s,Y_s)$ is the joint empirical distribution of the source.
Once trained on the source dataset, both $E_s(X_s)$ and $C(Z_s)$ will be maintained fixed during adaptation.

\subsection{Adversarial objectives}
To minimize the impact of the variation caused by the differences between source and target domains, we utilize the standard adversarial objective, $L_{\rm adv}$, to learn the discriminator $D(\cdot)$ by maximizing
\begin{align}\label{eq:l_adv}
	\begin{aligned}
	L_{\rm adv} = & \ {\mathbb E}_{x\sim p(X_s)} \ \log D(E_s(x)) \\
	+ & \ {\mathbb E}_{x\sim p(X_t)} \ \log( 1-D(E_t(x)) ) \,,
	\end{aligned}
\end{align}
where $p(X_s)$ and $p(X_t)$ are the marginal empirical distributions for source and target, respectively.
For the generator, we separately maximize over the target encoder via
\begin{equation}\label{eq:l_gen}
	L_{\rm enc}={\mathbb E}_{x\sim p(X_t)} \ \log( D(E_t(x)) ) \,,
\end{equation}
where we have inverted the labels relative to \eqref{eq:l_adv} as in \cite{goodfellow2014generative}, which has the same properties of the original min-max (adversarial) loss used in GAN but results in stronger gradients for the target encoder.

\subsection{Clustering objective}
Assuming that in latent space we have as many clusters, $K$, as distinct labels in the source domain, so $K=K_s$, we denote their centroids as $Z_c$, for $c=1,\ldots,K_s$.
Borrowing from Deep Unsupervised Embedding \cite{DEC}, we minimize the following Kullback-Leibler (KL) divergence
\begin{equation}\label{eq:l_dec}
	L_{\rm dec} = KL(P||Q) = \sum_{i=1}^N\sum_{c=1}^{K_s} p_{ic}\log\frac{p_{ic}}{q_{ic}} \,,
\end{equation}
where $Q$ and $P$ are the soft assignment and auxiliary distributions respectively.
For $Q$, we use soft-assignments from a mixture of Student's $t$ distributions with $\alpha$ degrees of freedom \cite{peel2000robust}, written as
\begin{equation*}
	q_{ic}=\frac{(1+||Z_i-Z_c||^2/\alpha)^{ -\frac{\alpha+1}{2} }}{ \sum_{ c^\prime }(1+||Z_i-Z_{c^\prime}||^2/\alpha)^{-\frac{\alpha+1}{2}} } \,,
\end{equation*}
from which $q_{ic}$ approximates the probability of instance \emph{i} being assigned to cluster \emph{c}.
We set $\alpha=1$ in all experiments.

For $P$, the auxiliary distribution, we encourage cluster tightness by raising $q_{ic}$ to a power of 2 and normalizing accordingly, so
\begin{equation}\label{eq:p_ic}
	p_{ic}=\frac{f_c^{-1}q_{ic}^2}{\sum_{c^\prime} f_{c^\prime}^{-1} q_{ic^\prime}^2} \,,
\end{equation}
where $f_c=\sum_i q_{ic}$.
Note that \eqref{eq:p_ic} naturally results in a self-reinforcement mechanism that encourages latent features $Z_t$ to lie closer to the centroids, $\{Z_c\}_{c=1}^{K_s}$, of the mixture distribution.
In the experiments, for the balanced and imbalanced cases, the $K_s$ cluster centroids, $Z_c$, are initialized to the mean of the latent representations of target instances that are predicted as class $c$ by $C(\cdot)$, for $c=1,\ldots,K_s$, so that each cluster is identified with one of the labels in the source.

\subsection{Cluster dissimilarity objective}
One limitation of the clustering loss in~\eqref{eq:l_dec} is that although it encourages clusters to be tight, it does not explicitly encourages them to be pure (consisting of members of the same class).
Moreover, in some cases it may result in domain collapse, \emph{i.e.}, clusters of distinct classes being located arbitrarily close.
To avoid these issues, we seek to match at the cluster level by encouraging that instances from different clusters are predicted as different classes.
So motivated, we define $A=[C(Z_1) \ \ldots \ C(Z_{K_s})]$ as the $K_s\times K_s$ matrix whose columns contain the distribution of label predictions for the $K_s$ centroids, $Z_c$, using classifier $C(Z_c)$.
Then define the cluster dissimilarity objective, $L_{\rm dis}$, to be minimized as
\begin{equation}\label{eq:l_dis}
	L_{\rm dis}=||A^\top A-I||_F \,,
\end{equation}
where $||\cdot||_F$ is the Frobenius norm.
Under~\eqref{eq:l_dis} entries of $A^\top A$ contain similarities between the class membership probabilities for all pairs of cluster centroids.
Note that diagonal entries of $A^\top A$ will encourage columns of $A$ to have unit norm.
As a result, columns of $A$ which represent probability vectors (positive and summing up to one) will be encouraged to become one-hot vectors.
By minimizing~\eqref{eq:l_dis} we encourage that the $K_s$ predicted class membership probability vectors are different and close to one-hot-vectors, in which case, each centroid will tend to be assigned to a different class with high probability.
For the implementation, we can write
\begin{equation}
	L_{\rm dis}=(\sum_c\sum_{j\ne i} (a_c^\top a_j)^2 )^{\frac{1}{2}} \,,
\end{equation}
where $a_c=C(Z_c)$ is a column of $A$ and compared to~\eqref{eq:l_dis}, we have excluded diagonal elements of $A^\top A$.
We found empirically that excluding the diagonal terms stabilizes training, thus preferred in the experiments.

\begin{algorithm}[t!]
	\caption{Training with SGD.}
	\label{alg:alg}
	\begin{algorithmic}
	\STATE Let \{$\theta_{E_s}$, $\theta_{E_t}$, $\theta_{C}$, $\theta_{D}$,$Z_c$\} be the parameters for each model component.
	\STATE {\bfseries Input:} 
	\STATE \hspace{3mm} Source and target data: \{$X_s$, $Y_s$\}, $X_t$
	\STATE \hspace{3mm} Learning rates \{$\gamma_{\rm adv}$, $\gamma_{\rm enc}$, $\gamma_{\rm dec}$, $\gamma_{\rm dis}$\}
	\STATE \hspace{3mm} Batch size $M$
	\STATE \hspace{3mm} Pre-training steps: $I_{\rm adv}$
	\STATE {\bfseries Output:}
	\STATE \hspace{3mm} Target encoder: $E_t(\cdot)$
	\STATE \hspace{3mm} Classifier: $C(\cdot)$
	\STATE Training a source model, $E_s(\cdot)$ and $C(\cdot)$, with $L_{\rm cla}$
	\STATE $i=0$
	\WHILE {not converge}
	\STATE Draw random minibatch $\{X_s\}_{s=1}^M$, $\{X_t\}_{t=1}^M$
	\IF{$i>I_{\rm adv}$}
	\STATE $\theta_{E_t} = \theta_{E_t} - \gamma_{\rm dec} \nabla_{\theta_{E_t}} L_{\rm dec}$
	\STATE $Z_c = Z_c - \gamma_{\rm dec} \nabla_{Z_c} L_{\rm dec}$
	\STATE $Z_c = Z_c - \gamma_{\rm dis} \nabla_{Z_c} L_{\rm dis}$
	\ENDIF
	\STATE $\theta_{D} = \theta_{D} - \gamma_{\rm adv} \nabla_{\theta_{D}} L_{\rm adv}$
	\STATE $\theta_{E_t} = \theta_{E_t} - \gamma_{\rm enc} \nabla_{\theta_{E_t}} L_{\rm enc}$
	\STATE $i = i + 1$
	\ENDWHILE
\end{algorithmic}
\end{algorithm}

\subsection{Complete objective}
The proposed robust unsupervised adaptation approach proceeds by first optimizing $L_{\rm cla}$ in~\eqref{eq:l_cla} on the source data, $\{X_s,Y_s\}_{s=1}^{N_s}$.
Then, with fixed source encoder and classifier, $E_s(X_s)$ and $C(Z_s)$, respectively, we will perform domain adaptation by updating in sequence the discriminator, target encoder and cluster centroids, $\{Z_c\}_{c=1}^{K_s}$, using the following complete objective: $L=L_{\rm adv}+L_{\rm enc}+L_{\rm dec}+L_{\rm dis}$, as in \eqref{eq:l_adv}, \eqref{eq:l_gen}, \eqref{eq:l_dec} and \eqref{eq:l_dis}.
Instead of specifying parameters in the complete objective to balance the different losses, we set different learning rates for each loss component as shown in Algorithm~\ref{alg:alg}.
In our experiments, $\gamma_{\rm enc}$ and $\gamma_{\rm adv}$ were set to the values specified originally in ADDA \cite{ADDA} and further set $\gamma_{\rm dis}=2\gamma_{\rm dec}$.
We will show that the model is fairly insensitive to the choice of $\gamma_{\rm dec}$.
Note that during the adaptation, the source labels are not needed and the source instances are only used to update the discriminator.
We found empirically that is beneficial to train the model with only the adversarial objectives for several iterations, ($I_{\rm adv}=0\sim150$ in the experiments), to provide a good initialization point for the clustering loss.

\begin{figure}[t!]
	\centering
	\begin{subfigure}[Before]{
	\includegraphics[height=2.6cm, width=3.5cm]{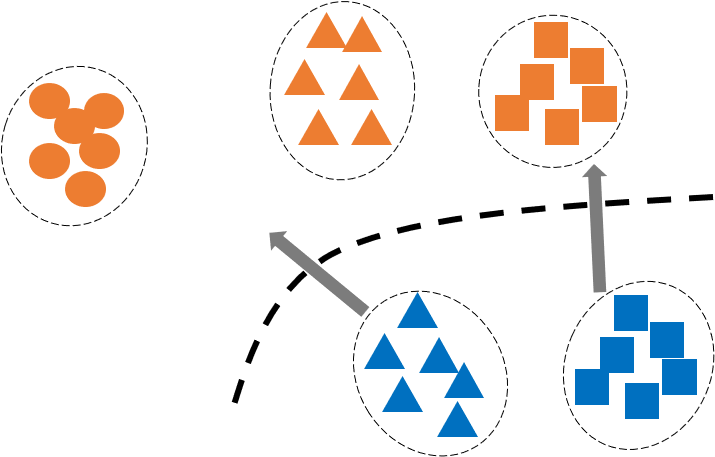}
	}
	\end{subfigure}
	\qquad
	\begin{subfigure}[After]{
	\includegraphics[height=2.6cm, width=3.5cm]{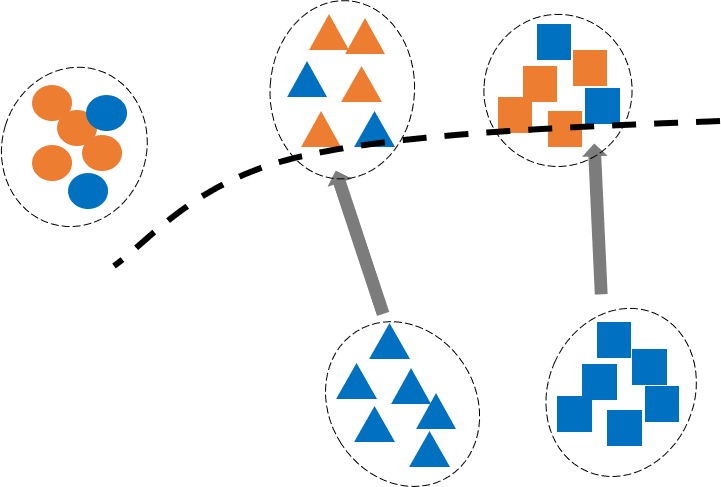}
	}
	\end{subfigure}
	\caption{Features before and after augmentation of the target set. The orange and blue items represent the source and target features. The dashed line is the boundary separating source and target learned by the discriminator. The arrows illustrate the direction of the adaptation. After the augmentation of the target inputs, the discriminator will set a softer boundary for the target domain. Since the target instances tend to transfer perpendicularly to the boundary via gradient updates, this modification can reduce negative transfer effects from the target toward outlier source classes.}
	\label{fg:partial}
\end{figure}

\subsection{Partial domain adaptation}\label{sc:partial}
In order to initialize the cluster centroids, we need to first specify the number of clusters, $K$.
In the setting considered above we simply let $K=K_s$, which is reasonable assuming source and target domains share the same label space.
However, this is not viable in the partial setting because the true number of target classes, $K_t$, is unknown.
If we set $K_t<K<K_s$ at least one of the clusters will be assigned to a label present in the source but not the target.
Alternatively, if we set $K<K_t<K_s$, at least one of the target classes will not be assigned.
Therefore, it seems cumbersome to guess the number of target classes.

Instead of attempting to estimate $K_t$, we propose a simple strategy consisting on \emph{augmenting} the target set with a portion of source data as shown in Figure~\ref{fg:partial}.
Specifically, when drawing a minibatch from the target to update the parameters of the model, $\{\theta_{E_t},\theta_D,Z_c\}$ in Algorithm~\ref{alg:alg}, we augment it with samples from the source, \emph{e.g.}, 50\% of the minibatch is drawn from the target and 50\% from the source (without using labels).
In this way, we can ensure that the augmented target has samples from all classes from the source.
As a result, the partial domain adaptation task is converted to a \emph{pseudo} imbalanced domain adaptation problem such that $K=K_s=K_t$, which is appropriate for our formulation.
It is worth noting that when augmenting the target set with source instances we do not use their label information.
Further, in Figure~\ref{fg:partial} we see that augmenting the target data with (a subset of) the source has the potential benefit of making it more difficult for the discriminator to distinguish source and target instances, thus resulting on a softer discriminator boundary for the target that can help reduce the effects of negative transfer.

In this setting, the cluster centroids are initialized using the source data rather than the combination of source and target instances.
In a real setting where we have no insight of whether the target contains all or a subset or the source classes, we can first treat it as imbalanced domain adaptation, then switch to partial domain adaptation if noticeable negative transfer is observed, \emph{e.g.}, by inspecting the $t$-SNE embedding of the learned latent representations.

\section{Related Work}
Unsupervised domain adaptation in the balanced setting has been extensively studied.
The general idea is to match the source and target marginal distributions directly or indirectly.
The latter, by matching their latent representations,
\cite{CORAL, deepCORAL} proposed to match the moments of different encoding layers of the latent representations.
This approach is easy to implement and achieves competitive results on several benchmarks.
\cite{DAN, DomainConfusion} used the Maximum Mean Discrepancy (MMD) framework to implicitly measure the distance between source and target distributions.
Specifically, they match the kernel embeddings of both distributions by minimizing their MMD.
Further, \cite{RTN} improved the MMD-based approaches by allowing separate classifiers for source and target domains.

Driven by the increasing popularity of the Generative Adversarial Networks (GANs) \cite{GANs}, recent adaptation methods resort to matching the distributions in an adversarial manner.
\cite{DAN, ADDA} added a discriminator to the latent representation to distinguish features from different domains, while the feature encoders are trained to mislead the domain discriminator so it cannot find an effective boundary that distinguishes between source and target instances.
The domain discriminator and feature encoders are trained adversarially as a min-max objective.
Inspired by \cite{ben2010theory}, \cite{ClassifierDiscrepancy} utilized the classifier discrepancy to detect target samples that are distant from the source.
Instead of using a discriminator, they proposed to adversarially maximize the discrepancy between two source classifiers, while training a feature encoder to reduce the inconsistency of their predictions.

The approaches described above rely on the assumption that source and target share the same label domain and distribution.
This assumption limits their applicability to situations where these are violated, \emph{i.e.}, imbalanced or partial domain adaptation scenarios.
\cite{learning2cluster} utilized the pairwise similarity information from the source to regularize the implicit clustering of the target domain and thus it has the potential to be used for the imbalanced scenario.
However, their clustering on target domain is determined from the source only, thus it does not benefit from the local information provided by the target.
\cite{SAN, PADA} introduced the concept of partial domain adaptation, in which target classes are assumed to be a subset of the source domain.
They reduce the effect of negative transfer by selecting out classes not present in the target, however, their approaches are only moderate when the source and target label domains are the same.
In our approach, we first transform the partial scenario into a special imbalanced setting via target domain augmentation, then we perform domain adaptation with our clustering-based objective without further changes.

\section{Experiments} \label{sc:experiments}
We conduct experiments on three domain adaptation benchmark datasets: the digits datasets, Office31 and VisDA2017.
The results below demonstrate that our method is robust to the difference in source and target label distributions by producing state-of-the-art classification performance in all of the three scenarios considered.
To quantify the impact of the newly introduced dissimilarity loss, we perform experiments with and without it, denoted as Ours and Ours (no $L_{dis}$), respectively.
In order to validate Figure~\ref{fg:partial}, we also define ADDA-mix, which corresponds to standard ADDA with argumented target inputs as described in Section~\ref{sc:partial}.
The mixture rate for source and target is 1:1 for all partial domain adaptation experiments.

\subsection{Datasets}
\textbf{The digits datasets:} We consider three digits datasets with varying difficulties: MNIST, SVHN and USPS, each containing 10 classes for digits 0-9.
The encoder architecture for the digits images is the modified LeNet from \cite{ADDA}.
For the domain classification, the adversarial discriminator consists of 3 fully connected layers with 500 hidden units for the first two layers and 2 for the output.
All images are converted to grayscale and rescaled to $28\times28$ pixels.
We consider three directions of transfer: SVHN$\rightarrow$MNIST, USPS$\rightarrow$MNIST and MNIST$\rightarrow$USPS.

\vspace{2mm}
\textbf{Office31:} This is a standard benchmark for domain adaptation widely used in computer vision, it consists of 4652 images from 31 classes.
These images are collected from three distinct domains: Amazon (A), which contains images downloaded from amazon.com, Webcam (W) and DSLR (D), which contain images taken by a web and a digital SLR camera, respectively, with different background settings.
This is a relatively difficult dataset since the Webcam and DSLR contains very small amount of images, \emph{i.e.}, less than 10 for some classes, which may easily lead to overfitting during the adaptation process.

In the experiments, we consider all the six directions of adaptation: A$\rightarrow$W, A$\rightarrow$D, W$\rightarrow$A, W$\rightarrow$D, D$\rightarrow$W and D$\rightarrow$A.
The architecture of the encoder for images in Office31 is a Resnet-50 \cite{ResNet} pre-trained on ImageNet.
All the images are first resized to $256\times256$ pixels RGB images, then random cropped during training and central cropped during testing into $224\times224$ RGB images.
Due to the small size of Office31, we approach the task as fully transductive, where all labeled instances from the source and all unlabeled instances from the target are used during training and adaptation. This is the same for the experiments in \cite{DAN,DANN,ADDA}.

\begin{figure}[t!]
	\vskip -0.2in
	\centering
	\begin{subfigure}[ADDA]{
		\includegraphics[height=3.7cm,trim={20mm 0 16mm 0},clip]{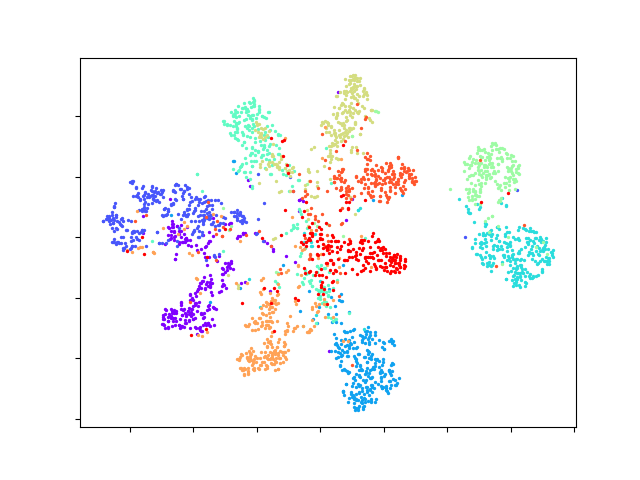}
		}
	\end{subfigure}
	\begin{subfigure}[Ours]{
		\includegraphics[height=3.7cm,trim={20mm 0 16mm 0},clip]{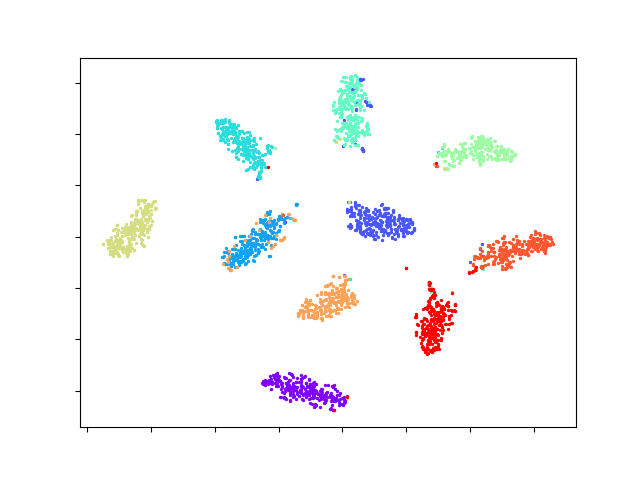}
		}
	\end{subfigure}
	\caption{$t$-SNE plot of the feature domains for SVHN$\rightarrow$MNIST.}
	\label{fg:svhn_mnist}
\end{figure}

\vspace{2mm}
\textbf{VisDA2017}: This is a dataset for the Visual Domain Adaptation Challange from synthetic 2D renderings of 3D models to real images.
It consists of 12 classes shared by both domains, each with a very large number of instances.
We use ResNet-50 as the source and target encoders.
Complementary to Office31, this dataset will validate the performance of our method on large-scale datasets.

\begin{figure*}[t]
	\centering
	\begin{subfigure}[Source Model]{
		\includegraphics[height=3.3cm,trim={20mm 0 16mm 0},clip]{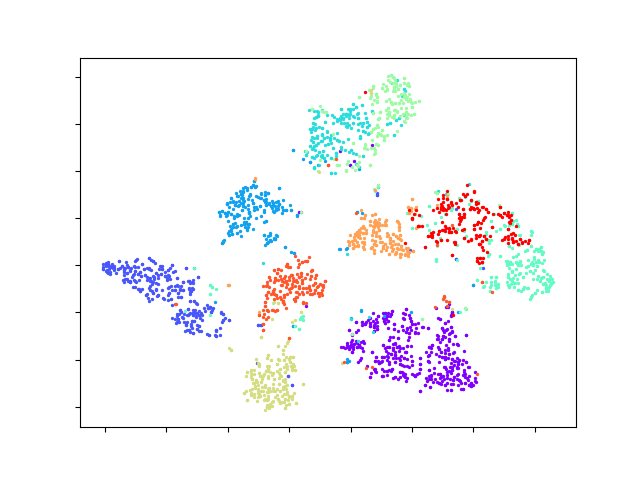}
		\label{fig:first}}
	\end{subfigure}
	\qquad
	\begin{subfigure}[ADDA]{
		\includegraphics[height=3.3cm,trim={20mm 0 16mm 0},clip]{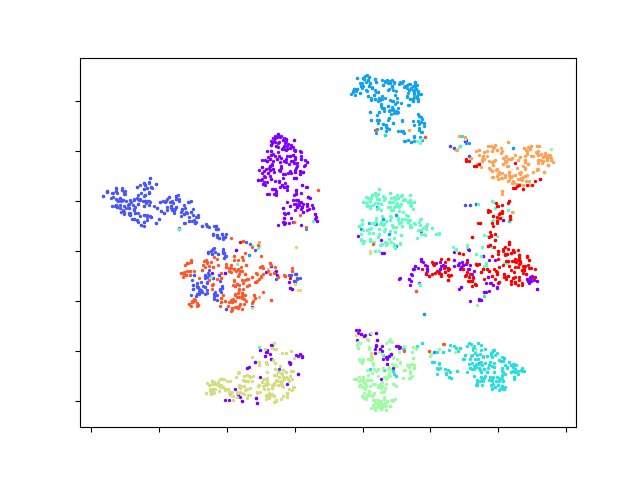}
		\label{fig:second}}
	\end{subfigure}
	\qquad
	\begin{subfigure}[DANN]{
		\includegraphics[height=3.3cm,trim={20mm 0 16mm 0},clip]{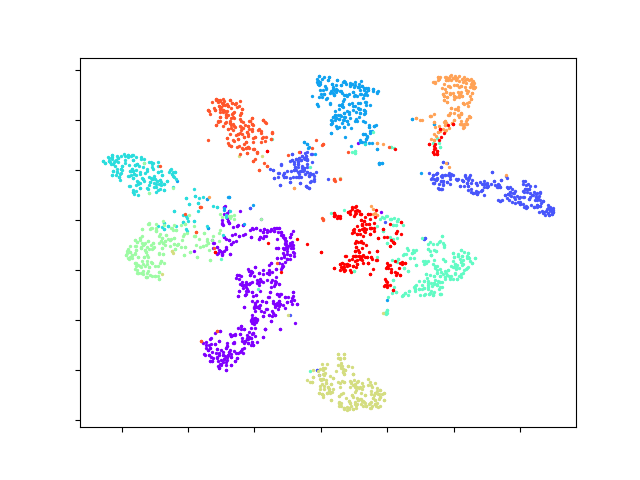}
		\label{fig:third}}
	\end{subfigure}
	\qquad
	\begin{subfigure}[Ours]{
		\includegraphics[height=3.3cm,trim={20mm 0 16mm 0},clip]{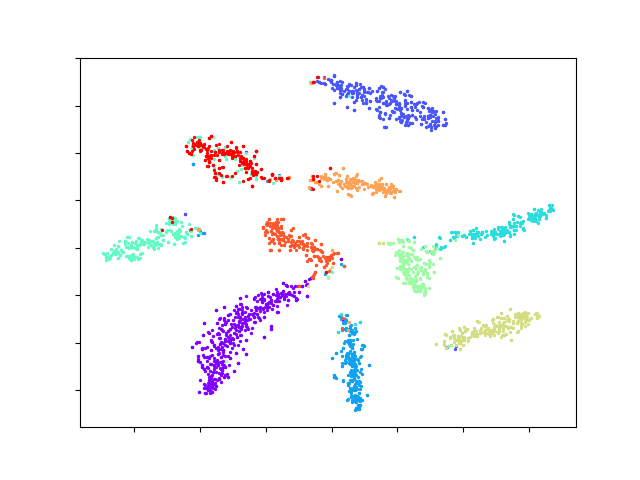}
		\label{fig:forth}}
	\end{subfigure}
	\caption{$t$-SNE embeddings of the target representations for MNIST$\rightarrow$USPS in the imbalanced setting.}
	\label{fg:tsne_imbalanced}
	\vskip 0.1in
\end{figure*}

\subsection{Balanced domain adaptation} \label{sc:balanced}
\textbf{The digit datasets:} Experiments are conducted with all the 10 digits in the balanced setting.
Results are shown in Table~\ref{tb:balanced}.
Our method outperforms all the other baselines in all three directions, which demonstrates the effectiveness of our method in the standard balanced setting with moderate to large datasets.
Figure~\ref{fg:svhn_mnist} shows the $t$-SNE embeddings of the target representation for SVHN$\rightarrow$MNIST, from which it can be observed that results from ADDA are more entangled, while those from ours are more concentrated and discriminative, thanks to the imposed clustering structure.
Further, if we do pure DEC clustering on MNIST without marginal domain matching, \emph{i.e}, without $L_{adv}$ and $L_{enc}$, the resulting clustering accuracy according to \cite{DEC} is 85\%, which is substantially lower than the accuracies with domain matching from SVHN$\rightarrow$MNIST and USPS$\rightarrow$MNIST (see Table~\ref{tb:balanced}).
These results highlight the benefits of jointly doing marginal domain matching and discriminative clustering.
%

\begin{table}[t!]
	\caption{Balanced domain adaptation on the digits datasets.}
	\label{tb:balanced}
	\centering
	\vskip -0.05in
	\begin{adjustbox}{tabular=lcccr,width=\columnwidth} 
      \\ \toprule
	Method & SVHN$\rightarrow$MNIST & USPS$\rightarrow$MNIST &MNIST$\rightarrow$USPS \\
	\midrule
	Source    & 0.598 & 0.634 & 0.771 \\
	DANN\cite{DANN}  & 0.746 & 0.909 & 0.880 \\
	ADDA\cite{ADDA}  & 0.760 & 0.901 & 0.894\\
	DIFA\cite{DIFA}    & 0.897 & 0.897 & 0.923 \\
      MCD\cite{ClassifierDiscrepancy} &0.962 & 0.941 & 0.942\\
      Adversarial Dropout\cite{adversarialdropout} & 0.950 & 0.931 & 0.932\\
	\midrule
    Ours (no $L_{dis}$) & \textbf{0.965}     & 0.976              & 0.943 \\
	Ours    & \textbf{0.965} & \textbf{0.979} & \textbf{0.952}        \\
	\bottomrule
	\end{adjustbox}
\end{table}

\begin{table}[t!]
	\caption{Balanced domain adaptation on Office31.}
	\label{tb:balanced_office}
	\vskip -0.05in
	\centering
	\begin{adjustbox}{tabular=ccccccc,width=\columnwidth}
	\\ \toprule
	Method & A$\rightarrow$ W  &  A$\rightarrow$D  &  W$\rightarrow$A  &  W$\rightarrow$D  &  D$\rightarrow$ W  &  D$\rightarrow$ A  \\
	\midrule
	Source   &   0.629             &     0.604               &         0.477           &         0.982            &    0.951             &     0.504  \\
	DAN\cite{DAN}     &    0.685             &      0.670              &         0.531           &         0.990            &    0.960             &     0.540  \\
	DANN\cite{DANN}   &    0.730             &       0.719              &         0.535           &        0.992            &    0.964             &     0.501  \\
	ADDA\cite{ADDA}   &   0.751             &      0.677               &         0.573           &         0.996            &   0.970              &     0.525  \\
	\midrule
       Ours (no $L_{dis}$)        & 0.787  & \textbf{0.743} & 0.535                   & 0.996                   &   \textbf{0.979}         & 0.532 \\
	Ours    &   \textbf{0.810}  & 0.727         & \textbf{0.595}      & \textbf{0.998}       &   \textbf{0.979} &  \textbf{0.553}  \\
	\bottomrule
	\end{adjustbox}
	\vskip -0.3in
\end{table}

\textbf{Office31:} In the case for balanced setting with Office31 data, we used all of the 31 classes in the three domains.
It is difficult to make the adaptation converge especially when the small domains (W and D) are used as target.
In our implementation, we set the initial learning rates for the discriminator and target encoder to 1e-3 and 1e-5.
The learning rates will be divided by 10 every 100 and 200 iterations respectively, with a batch size of 64.
Table~\ref{tb:balanced_office} shows the results for balanced domain adaptation for Office31.
Our method outperforms ADDA, DANN and DAN in all transfer directions, which validates its effectiveness on small datasets in the balanced adaptation scenario.

\begin{table*}[t]
	\caption{Imbalanced domain adaptation for MNIST$\rightarrow$USPS.}
	\label{tb:imbalanced}
	\vskip -0.1in
	\centering
	\begin{adjustbox}{tabular=cccccccccccc,width=1.8\columnwidth}
	\\ \toprule
	Method   & 0                  & 1                 & 2                  & 3                    & 4                 & 5                  & 6                & 7                  & 8                 & 9                  & Overall \\
	\midrule
	Source    & 0.816             & 0.962             & 0.874       & 0.663          &0.860              & 0.844             &0.888          &0.857            &0.747             &0.051             &0.771\\
	DANN\cite{DANN}     & 0.485            & 0.636            & 0.859        & \textbf{0.904} & \textbf{0.865} & 0.956            & 0.947       &0.590              &0.916            &0.548             &0.767\\
	ADDA\cite{ADDA}    & 0.493             & 0.640            & 0.874          & 0.874                & 0.795       & \textbf{0.963} &0.912         &0.939             &0.910            &0.825           &0.781\\
	\midrule
       Ours (no $L_{dis}$)  & 0.986           & 0.973        &   \textbf{0.980} &\textbf{0.934}&0.875      & 0.944              & 0.947            &\textbf{0.959}&\textbf{0.946}         &0.695              & 0.932 \\
	Ours      & \textbf{0.989} & \textbf{0.977} &   0.929  & 0.904          & 0.795            &\textbf{0.963}  &\textbf{0.965} & 0.932       &\textbf{0.964} &\textbf{0.876}  &\textbf{0.935} \\
	\bottomrule
	\end{adjustbox}
	\vskip -0.2in
\end{table*}

\subsection{Imbalanced domain adaptation}
We conduct an experiment for imbalanced domain adaptation on the MNIST$\rightarrow$USPS by manually sampling an imbalanced target domain.
From 0 to 9, the ratio of classes linearly decreases from 1 to 0.3 on USPS (target) and is kept uniform for the MNIST (source).
This means that, for the sampled USPS, if there are 10 images of 0s, then there are only 3 images for 9s.
The experiment is meant to illustrate the effects of negative transfer in the imbalanced setting and ability of our method to alleviate the negative transfer caused by the class imbalance.

Table~\ref{tb:imbalanced} shows accuracies on the target domain (USPS) for each class.
For ADDA and DANN, we should note that, although there is no obvious difference on the over all accuracies compared with the source, the large classes (0 and 1) are degraded during the adaptation due to the negative transfer illustrated in Figure~\ref{fg:negtrans}.
On the contrary, our method is robust against the imbalance and results in very high accuracies for most of the classes, especially 0s and 1s.
Further, is worth noting that the average accuracy of our method is only decreased by 0.017\% compared to the balanced scenario shown in Table~\ref{tb:balanced}.

The target representations for the four models are plotted in Figure~\ref{fg:tsne_imbalanced}.
For ADDA and DANN, it is clearly shown that the large classes, \emph{e.g.}, purple and dark blue for 0s and 1s, are negatively transferred toward other smaller classes when compared with the source model, while the target representation from our approach is more discriminative and better clustered.

\subsection{Partial domain adaptation}
\textbf{Office31:} We select the 10 classes shared by Office31 and Caltech-256 as our target labels.
For each direction of adaptation, we use all the images of these 10 classes in the target split as the target domain (denoted as A10, W10, D10), and images from all the 31 classes in the source split as the source domain (denoted as A31, W31, D31).
As described in Section~\ref{sc:partial}, we first convert the partial domain adaptation into a pseudo imbalanced setting, by augmenting the target with (unlabeled) source data, then normal domain adaptation is conducted as before.
In this setting, the cluster centroids are initialized using the source data instead of all source and target instances. Also we set $I_{adv} = 0$ in Algorithm 1, since training with raw adversarial objectives is likely to degrade the performance.

The results on Office31 are presented in Table~\ref{tb:office_imbalanced}.
SAN \cite{SAN} and PADA \cite{PADA} are two of the first approaches specifically designed for partial domain adaptation.
The results suggest that our method can outperformance SAN by a large margin and is competitive to PADA, which was proposed very recently.
Combined with the experiments on Office 31 in the balanced setting, these experiments validate the robustness of our method in learning discriminative target representations from small target datasets such as Office31.


\begin{table}[h!]
	\caption{Partial domain adaptation on VisDA2017.}
	\label{tb:partial_vis}
	\vskip -0.1in
	\centering
	\begin{adjustbox}{tabular=ccc,width=0.8\columnwidth}
	\\ \toprule
	Method & Syn-12$\rightarrow$Real-6& Real-12$\rightarrow$Syn6 \\
	\midrule
	Source    & 0.421            & 0.568\\
	DANN\cite{DANN}   &   0.327           &  0.605\\
	ADDA\cite{ADDA}  & 0.545              & 0.562\\
      ADDA-mix              & 0.543             &0.605\\
	\midrule
	PADA\cite{PADA}    & 0.535   & 0.765\\
      Ours (no $L_{dis}$)                 & 0.682    & 0.709 \\
	Ours & \textbf{0.700}  & \textbf{0.846}\\
	\bottomrule
	\end{adjustbox}
	\vskip -0.3in
\end{table}

\begin{table*}[t]
	\caption{Partial domain adaptation on Office31.}
	\label{tb:office_imbalanced}
	\vskip -0.1in
	\centering
	\begin{adjustbox}{tabular=ccccccc,width=1.4\columnwidth}
	\\ \toprule
	Method & A31$\rightarrow$ W10  &  A31$\rightarrow$D10  &  W31$\rightarrow$A10  &  W31$\rightarrow$D10  &  D31$\rightarrow$ W10  &  D31$\rightarrow$ A10  \\
	\midrule
	Source   &   0.664                       &     0.701                       &          0.691                  &        0.968                   &      0.980                      &     0.690  \\
	DANN\cite{DANN}   &    0.498                       &     0.529                      &          0.496                   &        0.624                   &      0.314                      &     0.468  \\
	ADDA\cite{ADDA}   &   0.593                        &     0.675                      &          0.727                   &        0.764                   &      0.705                      &     0.686  \\
      ADDA-mix             &    0.610                       &      0.713                     &          0.722                   &        0.949                   &      0.905                      &      0.707  \\
	\midrule
	SAN\cite{SAN}      &   0.800                       &     0.813                       &         0.831                    & \textbf{1.000}             &       0.986                     &     0.806  \\
	PADA\cite{PADA}    &   \textbf{0.865}          &     0.822                       &   \textbf{0.954}              & \textbf{1.000}             &        0.993                    &     0.927   \\     
      Ours (no $L_{dis}$)                   & 0.834                        &     0.834                      & 0.944               & \textbf{1.000}                 &  0.902                         &   0.886  \\
	Ours     &    0.834                       & \textbf{0.847}               &        0.943                    & \textbf{1.000}             &   \textbf{0.997}            &    \textbf{0.928}  \\
	\bottomrule
	\end{adjustbox}
	\vskip -0.1in
\end{table*}

\textbf{VisDA2017:} The VisDA2017 dataset was originally used for the balanced domain adaptation setting with shared label domain and distribution.
Following \cite{PADA}, we only reserve images of the first 6 classes in alphabetic order in the target domain (REAL-6, SYN-6), and all the images of the 12 classes are kept in the source domain (REAL-12, SYN-12).
The data preprocessing and experiment setting are the same as above for Office31 except that use ResNet-50 with 12-dimensional output instead of 31.

The results for SYN12$\rightarrow$REAL6 and REAL12$\rightarrow$SYN6 are shown in Table~\ref{tb:partial_vis}.
We see that our method outperforms PADA and the source by a large margin, which further demonstrates the ability of our approach on large scale datasets in the partial setting.

\subsection{Ablation test}
In order to demonstrate empirically the impact of the dissimilarity loss on performance, we considered throughout all experiments an ablation test for our model with (Ours) and without (Ours (no $L_{\rm dis}$)) the dissimilarity loss.
Note also that our model without both clustering and dissimilarity losses reduces to standard ADDA and on the partial domain adaptation to ADDA-mix.
In general, we observed that our approach with and without dissimilarity loss consistently outperforms ADDA and ADDA-mix, while ADDA-mix is generally better than ADDA.
Further, the dissimilarity loss often results in performance gains relative to the model without it.
%
These results suggest that training with $L_{\rm dec}$ and $L_{\rm dis}$ indeed helps in producing more discriminative target feature spaces.
It also demonstrates that the augmented target in partial adaptation can reduce the negative transfer when the source and target label domains are different, as described in Section~\ref{sc:partial}.

\begin{figure}[t]
	\centering
	\includegraphics[width=0.8\columnwidth]{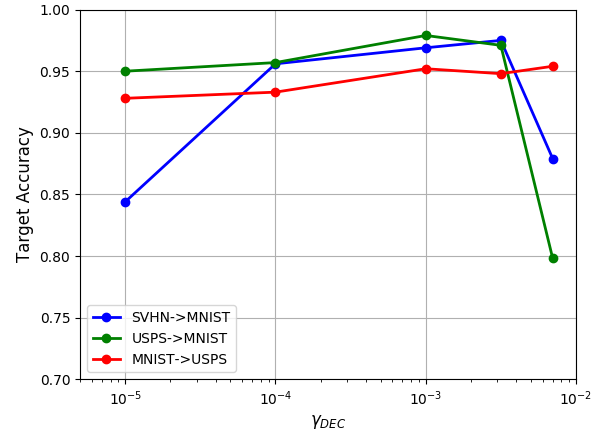}
	\caption{Accuracy on target testing data with different values of $\gamma_{DEC}$.}
	\label{fg:lr}
	\vskip -0.1in
\end{figure}

\subsection{Learning rates for clustering}
Our approach improves the discriminative ability of the target representation by encouraging target clustering via $L_{\rm dec}$ and $L_{\rm dis}$.
However, by doing so we introduce additional complexity and tuning requirements.
Therefore, we conduct a sensitive analysis for $\gamma_{\rm dec}$ and $\gamma_{\rm dis}$ on the digit datasets with settings similar to that of Section~\ref{sc:balanced}.

For this experiment, all other hyperparameters of the model are fixed and set to those in ADDA.
The relation between $\gamma_{\rm dec}$ and $\gamma_{\rm dis}$ is set to $\gamma_{\rm dis}=2\gamma_{\rm dec}$, as previously described.
We do so because the cluster centers are updated with both $L_{\rm dec}$ and $L_{\rm dis}$, but we seek for $L_{\rm dis}$ to dominate the update to promote class-dissimilar clusters and avoid domain collapse.
We consider all three adaptation directions, \emph{i.e.}, SVHN$\rightarrow$MNIST, USPS$\rightarrow$MNIST and MNIST$\rightarrow$USPS.
%

The resulting test accuracies for different values of $\gamma_{\rm dec}$ are shown in Figure~\ref{fg:lr}.
For MNIST$\rightarrow$USPS, we find that the target accuracy is not sensitive to the clustering learning rates on the testing range, \emph{i,e,}, $[1e-5,7e-3]$.
This suggests that, with small $\gamma_{\rm dec}$, though the algorithm can take longer to converge (results not shown), the model will converge to a reasonable optimal solution.
Alternatively, with large $\gamma_{\rm dec}$, the clustering strategy alone is able to produce a discriminative target representation.
On the other hand, for USPS$\rightarrow$MNIST and SVHN$\rightarrow$MNIST, the performance will drop dramatically when $\gamma_{\rm dec}$ is too large.
Nevertheless, we can still outperform ADDA by a large margin (see Table~\ref{tb:balanced}), as long as the selected learning rates are not too extreme.

\section{Conclusion}
We proposed a new domain adaptation method that extends the ability of existing adaptation approaches based on distribution matching to imbalanced and partial scenarios.
Our method improves the discrimination of target representations by simultaneously learning tightly clustered target embeddings and by encouraging each cluster to be assigned to a unique and different class from the source.
These criteria guarantee the robustness of the proposed method against differences between the source and target label distributions, which relaxes the common assumption that source and target share the same label domain and distribution.
We focused on three scenarios, namely, balanced domain adaptation, imbalanced domain adaptation and partial domain adaptation.
Experiments on several benchmark datasets demonstrated the effectiveness of the method on all the three scenarios, achieving state-of-the-art performances.

As future work, we are interested in extending our approach to semisupervised domain adaptation and segmentation adaptation, in which we will seek for pixels from the same segment in the target domain to be clustered in feature space and located nearby source features of the same class.
This is of interest particularly in medical imaging where segmentation is very expensive and time consuming.
Further, the clustering objective used in our approach relies on a reasonable initialization of the centroids.
Additional work is needed on more robust clustering procedures that are less sensitive to centroid pre-initialization.

\nocite{DEC}
\nocite{diss}
\nocite{DAN}
\nocite{DANN}
\nocite{ADDA}
\nocite{SAN}
\nocite{RTN}
\nocite{PADA}
\nocite{CORAL}
\nocite{deepCORAL}
\nocite{p2pDA}
\nocite{ResNet}
\nocite{shi2012information}
\nocite{learning2cluster}
\nocite{radford2015unsupervised}
\nocite{goodfellow2014generative}
\nocite{kingma2013auto}
\nocite{doersch2016tutorial}
\nocite{cui2017effect}
\nocite{UNet}
\nocite{patel2015visual}
\nocite{xie2018learning}
\nocite{openset}
\nocite{zhang2018importance}
\nocite{kang2018deep}
\nocite{chen2017cost}
\nocite{haeusser2017associative}
\nocite{DIFA}


\end{document}